\documentclass{article}

\PassOptionsToPackage{numbers}{natbib}
\usepackage[preprint]{neurips_2025}

\usepackage[utf8]{inputenc} % allow utf-8 input
\usepackage[T1]{fontenc}    % use 8-bit T1 fonts
\usepackage{url}            % simple URL typesetting
\usepackage{booktabs}       % professional-quality tables
\usepackage{nicefrac}       % compact symbols for 1/2, etc.
\usepackage{microtype}      % microtypography
\usepackage{graphics,graphicx,color}

\usepackage{algorithm}
\usepackage{algorithmic}
\usepackage{enumitem}

\usepackage{amsfonts}       % blackboard math symbols
\usepackage{amsmath}       % blackboard math symbols
\usepackage{amssymb}

\newcommand{\Figref}[1]{Figure~\ref{#1}}  % beginning of sentence
\newcommand{\figref}[1]{Fig.~\ref{#1}}    % somewhere

\newcommand{\tabref}[1]{Table~\ref{#1}}

\newcommand{\eqnref}[1]{Eq.~\ref{#1}} % Eq. (1)
\newcommand{\secref}[1]{Sec.~\ref{#1}} % Sec. 1

\newcommand{\Algref}[1]{Algorithm~\ref{#1}}

\usepackage{xspace}
\makeatletter
\DeclareRobustCommand\onedot{\futurelet\@let@token\@onedot}
\def\@onedot{\ifx\@let@token.\else.\null\fi\xspace}
\makeatother

\usepackage{xr-hyper}
\makeatletter
\newcommand*{\addFileDependency}[1]{% argument=file name and extension
  \typeout{(#1)}
  \@addtofilelist{#1}
  \IfFileExists{#1}{}{\typeout{No file #1.}}
}
\makeatother

\usepackage{xcolor}
\definecolor{ourorange}{HTML}{e19c24}
\definecolor{ourgreen}{HTML}{97b567}
\definecolor{ourred}{HTML}{ec6235}
\definecolor{ourblue}{HTML}{5e81b5}
\definecolor{ourgrey}{HTML}{919191}

\definecolor{myyellow}{HTML}{ffc800}

\usepackage{subcaption}

\DeclareMathAlphabet{\mathsfit}{\encodingdefault}{\sfdefault}{m}{sl}
\SetMathAlphabet{\mathsfit}{bold}{\encodingdefault}{\sfdefault}{bx}{n}

\def\gA{{\mathcal{A}}}

\def\gD{{\mathcal{D}}}

\def\gL{{\mathcal{L}}}

\def\gO{{\mathcal{O}}}

\def\gS{{\mathcal{S}}}

\def\sR{{\mathbb{R}}}

\newcommand{\E}{\ensuremath{\mathbb E}}           % Expectation
\def\subsubsubsection{\vskip
5pt{\noindent\normalsize\rm\raggedright}}
\newcommand{\papertitle}{\textcolor{myyellow}{R}obotic \textcolor{myyellow}{W}orld \textcolor{myyellow}{M}odel: A Neural Network Simulator for Robust Policy Optimization in Robotics\xspace}
\newcommand{\method}{RWM\xspace}
\newcommand{\methodfull}{Robotic World Model\xspace}
\newcommand{\policymethod}{MBPO-PPO\xspace}

\usepackage[utf8]{inputenc} % allow utf-8 input
\usepackage[T1]{fontenc}    % use 8-bit T1 fonts
\usepackage{hyperref}       % hyperlinks
\usepackage{url}            % simple URL typesetting
\usepackage{booktabs}       % professional-quality tables
\usepackage{amsfonts}       % blackboard math symbols
\usepackage{nicefrac}       % compact symbols for 1/2, etc.
\usepackage{microtype}      % microtypography
\usepackage{xcolor}         % colors

\title{\papertitle}

\author{%
  Chenhao Li \\
  ETH Zurich, Switzerland \\
  \texttt{chenhli@ethz.ch} \\
  \And
  Andreas Krause \\
  ETH Zurich, Switzerland \\
  \texttt{krausea@ethz.ch} \\
  \And
  Marco Hutter \\
  ETH Zurich, Switzerland \\
  \texttt{mahutter@ethz.ch} \\
}

\begin{document}

\maketitle

\begin{center}
\vspace{-2em}
\textcolor{myyellow}{\url{https://sites.google.com/view/roboticworldmodel}}
\end{center}

\begin{abstract}
Learning robust and generalizable world models is crucial for enabling efficient and scalable robotic control in real-world environments.
In this work, we introduce a novel framework for learning world models that accurately capture complex, partially observable, and stochastic dynamics.
The proposed method employs a dual-autoregressive mechanism and self-supervised training to achieve reliable long-horizon predictions without relying on domain-specific inductive biases, ensuring adaptability across diverse robotic tasks.
We further propose a policy optimization framework that leverages world models for efficient training in imagined environments and seamless deployment in real-world systems.
% Through extensive experiments, our approach consistently outperforms state-of-the-art methods, demonstrating superior prediction accuracy, robustness, and generalization across diverse robotic tasks.
% Notably, policies trained with our method are successfully deployed on ANYmal\,D hardware in a zero-shot transfer, achieving robust performance with minimal performance loss.
This work advances model-based reinforcement learning by addressing the challenges of long-horizon prediction, error accumulation, and sim-to-real transfer.
By providing a scalable and robust framework, the introduced methods pave the way for adaptive and efficient robotic systems in real-world applications.

\end{abstract}

\begin{figure}[h]
    \centering
    \includegraphics[width=0.95\linewidth]{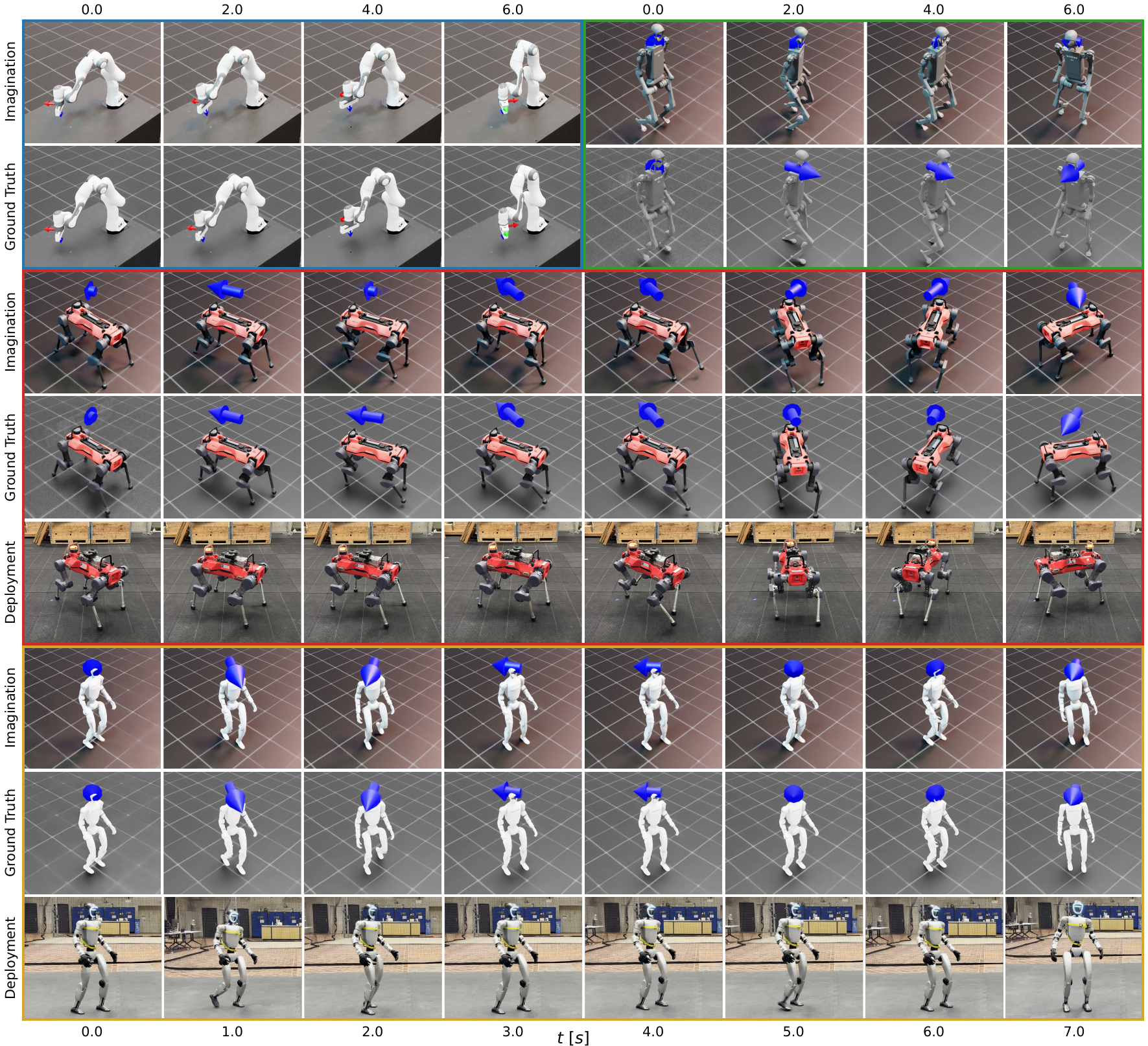}
    \caption{Autoregressive imagination, ground-truth simulation, and real-world deployment of \method. For each environment, the top row showcases the \method autoregressively predicting future trajectories in imagination. The second row visualizes the ground truth evolution in simulation. Specifically for the ANYmal\,D quadruped and Unitree\,G1 humanoid, the framework achieves robust policy optimization through \policymethod, enabling zero-shot deployment on hardware.}
    % across diverse robotic systems from the same initialization
    % The visualized coordinate and arrow markers denote the predicted and measured end-effector pose and base velocity, respectively.
    \label{fig:prediction_visualization}
\end{figure}

\section{Introduction}
Robotic systems have achieved remarkable advancements in recent years, driven by progress in reinforcement learning (RL)~\citep{haarnoja2018soft, schulman2017proximal} and control theory~\citep{nguyen2019review, todorov2012mujoco}.
% High-fidelity simulators have played a pivotal role in these developments, enabling safe and efficient policy training~\citep{todorov2012mujoco, makoviychuk2021isaac, mittal2023orbit}.
% However, the transfer of policies from simulation to real-world environments remains a significant challenge due to inevitable inaccuracies in modeling real-world dynamics~\citep{peng2018sim, hwangbo2019learning, li2023learning}.
% These inaccuracies lead to performance degradation when deploying learned policies on physical hardware, a problem commonly referred to as the sim-to-real gap.
A prevalent limitation in many approaches is the lack of adaptation and learning once the policy is deployed on the real system~\citep{tan2018sim, peng2020learning, li2023learning, li2023versatile}.
This results in underutilization of the valuable data generated during real-world interactions.
Robotic systems operating in dynamic and uncertain environments require the ability to continually adapt their behavior to new conditions~\citep{lee2020learning}.
The inability to exploit real-world experience for further learning restricts the system’s robustness and limits its ability to handle evolving scenarios effectively.
Truly intelligent robotic systems should operate efficiently and reliably using limited data, adapting to real-world conditions in a scalable manner~\citep{nagabandi2018neural, hafner2020mastering}.
While model-free RL algorithms such as Proximal Policy Optimization (PPO)~\citep{schulman2017proximal} and Soft Actor-Critic (SAC)~\citep{haarnoja2018soft} have demonstrated impressive results in simulation, their high interaction requirements make them impractical for real-world robotics.
Sample-efficient methods are therefore essential for leveraging the information in real-world data without extensive environment interactions~\citep{chua2018deep, janner2019trust}.

A promising solution is the use of predictive models of the environment, commonly referred to as world models~\citep{ha2018recurrent, hafner2019learning}.
World models simulate environment dynamics to enable planning and policy optimization, often referred to as \textit{learning in imagination}~\citep{sutton1991dyna}.
These models have shown success across diverse robotic domains, including manipulation~\citep{ebert2018visual, finn2016unsupervised}, navigation~\citep{hafner2020mastering}, and locomotion~\citep{nagabandi2018neural}.
However, developing reliable and generalizable world models poses unique challenges due to the complexity of real-world dynamics, including nonlinearities, stochasticity, and partial observability~\citep{wu2023daydreamer, song2024learning}.
Existing approaches often incorporate domain-specific inductive biases, such as structured state representations or hand-designed network architectures~\citep{yang2020data, sancaktar2022curious, li2024fld}, to improve model fidelity.
While effective, these methods are limited in their scalability and adaptability to novel environments or tasks.
In contrast, a general framework for learning world models without domain-specific assumptions has the potential to enhance generalization and applicability across a wide range of robotic systems and scenarios.

In this work, we present a novel approach for learning world models that emphasizes robustness and accuracy over long-horizon predictions.
Our method is designed to operate without handcrafted representations or specialized architectural biases, enabling broad applicability to diverse robotic tasks.
To evaluate the utility of these learned models, we further propose a policy optimization method using PPO and demonstrate successful deployment in both simulated and real-world environments.
To the best of our knowledge, this is the first framework to reliably train policies on a learned neural network simulator without any domain-specific knowledge and deploy them on physical hardware with minimal performance loss.

{\bf Our contributions} are summarized as follows:
\textbf{(i)} We introduce a novel network architecture and training framework that enables the learning of reliable world models capable of long autoregressive rollouts, a critical property for downstream planning and control.
\textbf{(ii)} We provide a comprehensive evaluation suite spanning diverse robotic tasks to benchmark our method. Comparative experiments with existing world model frameworks demonstrate the effectiveness of our approach.
\textbf{(iii)} We propose an efficient policy optimization framework that leverages the learned world models for continuous control and generalizes effectively to real-world scenarios with hardware experiments, including both quadruped and humanoid systems.

By addressing the challenges associated with learning world models, this work contributes toward bridging the gap between data-driven modeling and real-world deployment.
The proposed framework enhances the scalability, adaptability, and robustness of robotic systems, paving the way for broader adoption of model-based reinforcement learning in real-world applications.
Supplementary videos for this work are available on \url{https://sites.google.com/view/roboticworldmodel}.

\section{Related work}
\subsection{World Models for Robotics}

World models have emerged as a cornerstone in robotics for capturing system dynamics and enabling efficient planning and control through simulated trajectories.
% Early works like PILCO~\citep{deisenroth2011pilco} demonstrated exceptional sample efficiency by leveraging Gaussian processes, but their applicability to high-dimensional, nonlinear systems has been limited by scalability constraints.
% The advent of deep neural networks expanded the horizons of dynamics modeling, allowing complex robotic systems to be effectively modeled and controlled.
% For example, neural network dynamics models have been integrated with Model Predictive Control (MPC) to enhance adaptability in real-world settings~\citep{nagabandi2018neural}.
A prominent application of world models is in robotic control, where dynamics models are used to describe real-world dynamics for policy optimization~\citep{levine2016end}.
Extensions to vision-based tasks have been realized through visual foresight techniques~\citep{finn2016unsupervised, finn2017deep, ebert2018visual}, which learn visual dynamics for planning in high-dimensional sensory spaces.
Similar ideas are applied to train RL agents in such world models aiming to fully replicate real environment interactions~\citep{ha2018recurrent, alonso2024diffusion}.
These approaches underline the versatility of world models in tasks requiring rich perceptual inputs.

To improve the generalization of black-box neural network-based world models beyond the training distribution, many works incorporate known physics principles or state structures into model design, addressing potential limitations in control performance.
Examples include foot-placement dynamics~\citep{yang2020data}, object invariance~\citep{sancaktar2022curious}, granular media interactions~\citep{choi2023learning}, frequency domain parameterization~\citep{li2024fld}, rigid body dynamics~\citep{song2024learning}, and semi-structured Lagrangian dynamics models~\citep{levy2024learning}.
While these methods demonstrate impressive results, they often require strong domain knowledge and carefully crafted inductive biases, which can restrict their scalability and adaptability to diverse robotic applications.
Latent-space dynamics models offer an alternative by abstracting the state space into compact representations, enabling efficient long-horizon planning.
Deep Planning Network (PlaNet)~\citep{hafner2019learning} and its successor Dreamer~\citep{hafner2019dream, hafner2020mastering, hafner2023mastering} exemplify this trend, achieving state-of-the-art performance in continuous control and visual navigation tasks.
These frameworks have been extended to real-world robotics~\citep{wu2023daydreamer, bi2024sample}, demonstrating their potential in both simulation and hardware deployment.

\subsection{Model-Based Reinforcement Learning}

Model-Based Reinforcement Learning (MBRL) has emerged as a powerful approach to address the limitations of model-free reinforcement learning, particularly in scenarios where sample efficiency and safety are critical.
Unlike model-free methods, which learn policies directly from interactions with the environment, MBRL leverages a learned model of the environment to simulate interactions, enabling more efficient and safer policy learning.
One of the pioneering methods in MBRL is Probabilistic Ensembles with Trajectory Sampling (PETS), which uses an ensemble of probabilistic neural networks to model the environment dynamics~\citep{chua2018deep}.
% This ensemble approach captures the uncertainty in the environment, allowing for more robust decision-making.
Building on the idea of latent-space modeling, PlaNet leverages a latent dynamics model to plan directly in a learned latent space~\citep{hafner2019learning}.
% This approach enables long-horizon planning and is particularly effective in environments where direct planning in the raw state space would be computationally prohibitive.
% Similar ideas were also seen in previous attempts~\citep{watter2015embed, ha2018world}.
% PlaNet has been applied to control tasks in simulated robotic environments, such as continuous control tasks and visual-based navigation, where it significantly reduces the computational burden while maintaining high levels of performance.
Dreamer extends the concept by incorporating an actor-critic framework into the latent dynamics model, enabling the simultaneous learning of both the dynamics model and the policy~\citep{hafner2019dream, hafner2020mastering, hafner2023mastering}.
Variations on the architectural design also see success in improving generation capabilities of such latent dynamics models with autoregressive transformer~\citep{micheli2022transformers} and the stochastic nature of variational autoencoders~\citep{zhang2024storm}.
Recent advancements in this area include TD-MPC and TD-MPC2, which integrate model-based learning with MPC to achieve high-performance control in dynamic environments~\citep{hansen2022temporal, feng2023finetuning, hansen2023td}.
% TD-MPC leverages temporal difference learning to refine the model and improve the accuracy of long-term predictions~\citep{hansen2022temporal, feng2023finetuning}, while TD-MPC2 extends this approach by incorporating more sophisticated planning strategies to handle environments with high variability~\citep{hansen2023td}.

Recognizing the strengths of both model-based and model-free methods, several hybrid approaches have been developed to combine the sample efficiency of MBRL with the robustness of model-free reinforcement learning.
One notable example is Model-Based Policy Optimization (MBPO), which uses a model-based approach for planning and policy optimization but refines the policy using model-free updates~\citep{janner2019trust}.
It emphasizes selectively relying on the learned model when its predictions are accurate, thus mitigating the negative effects of model inaccuracies.
% This approach has proven effective in environments where model dynamics can be learned with reasonable accuracy, allowing for more efficient policy optimization.
Building on similar principles, Model-based Offline Policy Optimization (MOPO) extends the framework to the offline setting, where learning is conducted entirely from previously collected data without further environment interaction~\citep{yu2020mopo}.
% MOPO introduces uncertainty penalization in the learned model’s predictions to ensure conservative policy updates, effectively reducing the risk of overfitting to model errors.
% This approach is particularly valuable in scenarios where real-world data collection is costly or risky, such as in medical decision-making or autonomous driving.
% Model-Based Meta-Policy-Optimization (MB-MPO) advances MBPO by incorporating meta-learning techniques to improve the generalization of learned policies across different tasks~\citep{clavera2018model}.
% This method trains a meta-policy that can quickly adapt to new environments by leveraging a shared model, enhancing the applicability of MBPO in multi-task settings.
% This approach bridges the gap between model-based and model-free learning by allowing the policy to be fine-tuned with minimal interaction in new environments.
In contrast to using zeroth-order model-free reinforcement learning for policy optimization, first-order gradient-based optimization is used to improve policy learning~\citep{xu2022accelerated, georgiev2024adaptive}.
% These methods scale up the world models to capture intricate details of environment dynamics and directly apply gradient-based optimization to improve policy learning.
This allows for more efficient and precise policy updates, particularly in complex, high-dimensional environments, where accurate gradient information is crucial for performance.
Our framework extends MBPO by integrating it with PPO over extensive autoregressive rollouts, making it particularly effective for complex robotic control tasks.

\section{Approach}
\label{sec:approach}

\subsection{Reinforcement Learning and World Models}

We formulate the problem by modeling the environment as a Partially Observable Markov Decision Process (POMDP)~\citep{sutton2018reinforcement}, defined by the tuple $\left (\gS, \gA, \gO, T, R, O, \gamma \right)$, where $\gS$, $\gA$, and $\gO$ denote the state, action, and observation spaces, respectively.
The transition kernel $T: \gS \times \gA \to \gS$ captures the environment dynamics $p \left (s_{t+1} \mid s_t, a_t \right )$, while the reward function $R : \gS \times \gA \times \gS \to \sR$ maps transitions to scalar rewards.
Observations $o_t \in \gO$ are emitted according to probabilities $p \left (o_t \mid s_t \right )$, governed by the observation kernel $O: \mathcal{S} \to \mathcal{O}$.
The agent seeks to learn a policy $\pi_\theta: \gO \to \gA$ that maximizes the expected discounted return $\E_{\pi_\theta} \left[\sum_{t \geq 0} \gamma^t r_t\right]$, where $r_t$ is the reward at time $t$ and $\gamma \in \left [0, 1 \right ]$ is the discount factor.

World models~\citep{ha2018recurrent} approximate the environment dynamics and facilitate policy optimization by enabling simulated environment interactions in \textit{imagination}~\citep{sutton1991dyna}.
Training typically involves three iterative steps: (1) collect data from real environment interactions; (2) train the world model using the collected data; and (3) optimize the policy within the simulated environment produced by the world model.

Despite the success of existing frameworks in achieving tasks in simplified settings, their application to complex low-level robotic control remains a significant challenge.
To address this gap, we propose \methodfull (\method), a novel framework for learning robust world models in partially observable and dynamically complex environments.
\method builds on the core concept of world models but introduces architectural and training innovations that enable reliable long-horizon predictions, even in stochastic and partially observable settings.
By incorporating historical context and autoregressive training, \method addresses challenges such as error accumulation and partially observable and discontinuous dynamics, which are critical in real-world robotics applications.

\subsection{Self-supervised Autoregressive Training}
\label{sec:self-supervised_autoregressive_training}

To address the inherent complexity of partially observable environments, we propose a self-supervised autoregressive training framework as the backbone of \method.
This framework trains the world model $p_\phi$ to predict future observations by leveraging both historical observation-action sequences and its own predictions, ensuring robustness over extended rollouts.

The input to the world model consists of a sequence of observation-action pairs spanning $M$ historical steps.
At each time step $t$, the model predicts the distribution of the next observation $p \left (o_{t+1} \mid o_{t-M+1:t}, a_{t-M+1:t} \right )$.
Predictions are generated autoregressively: at each step, the predicted observation $o'_{t+1}$ is appended to the history and combined with the next action $a_{t+1}$ to serve as input for subsequent predictions.
This process is repeated over a prediction horizon of $N$ steps, producing a sequence of future predictions.
The predicted observation $k$ steps ahead can thus be written as
\begin{equation}
    o'_{t+k} \sim p_\phi \left ( \cdot \mid o_{t-M+k:t}, o'_{t+1:t+k-1}, a_{t-M+k:t+k-1} \right ).
    \label{eqn:autoregressive}
\end{equation}
A similar process is also applied to predict privileged information $c$, such as contacts, providing an additional learning objective that implicitly embeds critical information for accurate long-term predictions.
Such a training scheme introduces the model to the distribution it will encounter at test time, reducing the mismatch between training and inference distributions.
Overall, the model is optimized by minimizing the multi-step prediction error:
\begin{equation}
    \gL = \frac{1}{N} \sum_{k=1}^N \alpha^k \left [ L_o \left (o'_{t+k}, o_{t+k} \right ) + L_c \left (c'_{t+k}, c_{t+k} \right ) \right ],
    \label{eqn:loss}
\end{equation}
where $L_o$ and $L_c$ quantify the discrepancy between predicted and true observations and privileged information, and $\alpha$ denotes a decay factor.
This autoregressive training objective encourages the hidden states to encode representations that support accurate and reliable long-horizon predictions.

\begin{figure}
    \centering
    \begin{subfigure}{0.45\linewidth}
        \includegraphics[width=\linewidth]{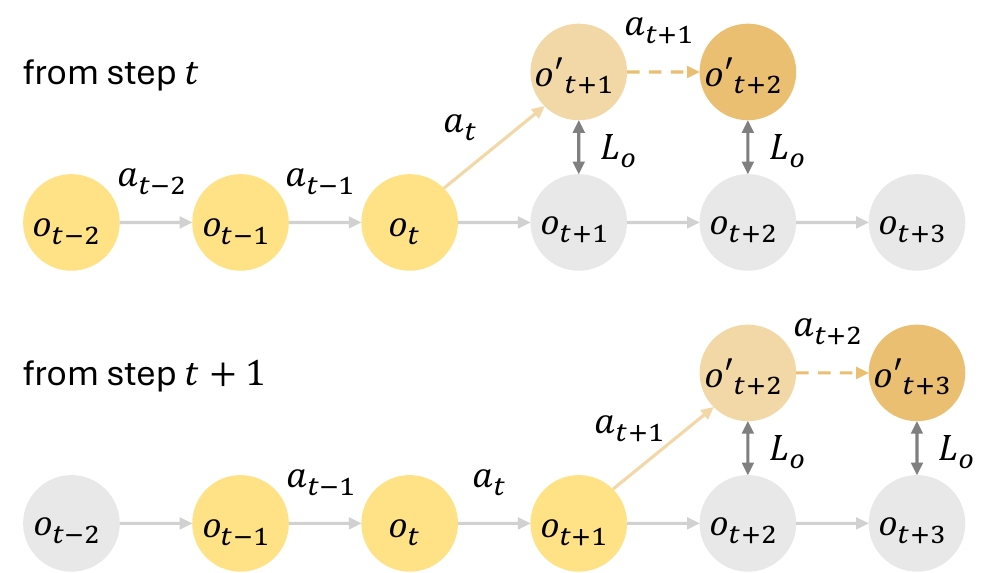}
        \caption{Autoregressive training.}
        \label{fig:autoregressive_training}
    \end{subfigure}
    \hspace{2em}
    \begin{subfigure}{0.45\linewidth}
        \includegraphics[width=\linewidth]{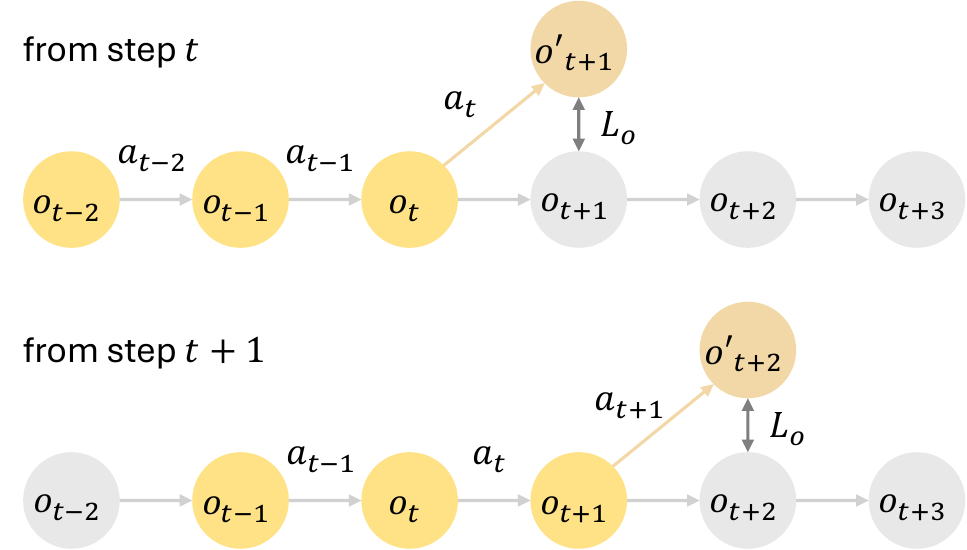}
        \caption{Teacher-forcing training.}
        \label{fig:teacher_forcing_training}
    \end{subfigure}
    
    \caption{Comparison of training paradigms for world models with an example of a history horizon $H = 3$. (a) Autoregressive training operates with an example of a forecast horizon $N = 2$, leveraging historical data and its own predictions for long-horizon robustness. The dashed arrows denote the sequential autoregressive prediction steps. (b) Teacher-forcing training can be viewed as a special case of autoregressive training with a forecast horizon $N = 1$, using ground truth observations for next-step predictions to optimize parallelization but limiting robustness to error accumulation.}
\end{figure}

Training data is constructed by sliding a window of size $M+N$ over collected trajectories, providing sufficient historical context for prediction targets.
To improve gradient propagation through autoregressive predictions, we apply reparameterization tricks to enable effective end-to-end optimization.
By incorporating historical observations, \method captures unobservable dynamics, addressing the challenges of partially observable and potentially discontinuous environments.
The autoregressive training mitigates error accumulation, a common issue in long-horizon predictions, and eliminates the need for handcrafted representations or domain-specific inductive biases, enhancing generalization across diverse tasks.
This process is illustrated in \figref{fig:autoregressive_training}, in contrast to the teacher-forcing pipeline in  \figref{fig:teacher_forcing_training}, which is commonly adopted to train many popular architectures~\citep{hafner2019dream, chen2021decision}.
Specifically, teacher-forcing can be viewed as a special case of autoregressive training with forecast horizon $N=1$, which boosts training with higher parallelization.

While the proposed autoregressive training framework can be applied to any network architecture, \method utilizes a GRU-based architecture for its ability to maintain long-term historical context while operating on low-dimensional inputs.
The network predicts the mean and standard deviation of a Gaussian distribution describing the next observation.
Our framework introduces a \textit{dual-autoregressive mechanism}: \textbf{(i)} \textit{Inner autoregression} updates GRU hidden states autoregressively after each historical step within the context horizon $M$.
\textbf{(ii)} \textit{Outer autoregression} feeds predicted observations from the forecast horizon $N$ back into the network.
This architecture, visualized in \figref{fig:dual-autoregressive}, ensures robustness to long-term dependencies and transitions, making \method suitable for complex robotics applications.

\subsection{Policy Optimization on Learned World Models}

Policy optimization in \method is conducted using the learned world model, following a framework inspired by Model-Based Policy Optimization (MBPO)~\citep{janner2019trust} and the Dyna algorithm~\citep{sutton1990integrated}.
During imagination, the actions are generated recursively by the policy $\pi_\theta$ conditioned on the observations predicted by the world model $p_\phi$, which is further conditioned on the previous predictions.
The actions at time $t+k$ can thus be written as
\begin{equation}
    a'_{t+k} \sim \pi_\theta \left (\cdot \mid o'_{t+k} \right),
    \label{eqn:policy}
\end{equation}
where $o'_{t+k}$ is drawn autoregressively according to \eqnref{eqn:autoregressive}.
Rewards are computed from imagined observations and privileged information.
The approach combines model-based imagination with model-free RL to achieve efficient and robust policy optimization, as outlined in \Algref{algorithm}.

\begin{algorithm}
    \caption{Policy optimization with \method}
    \label{algorithm}
    \begin{algorithmic}[1]
        \STATE Initialize policy $\pi_\theta$, world model $p_\phi$, and replay buffer $\gD$
        \FOR{learning iterations $ = 1,2,\dots$}
            \STATE Collect observation-action pairs in $\gD$ by interacting with the environment using $\pi_\theta$
            \STATE Update $p_\phi$ with autoregressive training using data sampled from $\gD$ according to \eqnref{eqn:loss}
            \STATE Initialize imagination agents with observations sampled from $\gD$
            \STATE Roll out imagination trajectories using $\pi_\theta$ and $p_\phi$ for $T$ steps according to \eqnref{eqn:policy}
            \STATE Update $\pi_\theta$ using PPO or another reinforcement learning algorithm
        \ENDFOR
    \end{algorithmic}
\end{algorithm}

The replay buffer $\gD$ aggregates real environment interactions collected by a single agent.
The world model $p_\phi$ is trained on this data following the autoregressive scheme described in \secref{sec:self-supervised_autoregressive_training}.
Imagination agents are initialized from samples in $\gD$ and simulate trajectories using the world model for $T$ steps, enabling policy updates through a reinforcement learning algorithm.
The training diagram is visualized in \figref{fig:mbpo}.

While PPO is known for its strong performance in robotic tasks, training it on learned world models poses unique challenges.
Model inaccuracies can be exploited during policy learning, leading to discrepancies between the imagined and true dynamics.
This issue is exacerbated by the extended autoregressive rollouts required for PPO, which compound prediction errors.
We denote this policy optimization method by \policymethod.
Despite these challenges, \method demonstrates its robustness by successfully optimizing policies over a hundred autoregressive steps with \policymethod, far exceeding the capabilities of existing frameworks such as MBPO~\citep{janner2019trust}, Dreamer~\citep{hafner2019dream, hafner2020mastering, hafner2023mastering}, or TD-MPC~\citep{hansen2022temporal, hansen2023td}.
This result underscores the accuracy and stability of the proposed training method and its ability to synthesize policies deployable on hardware.
% By using \policymethod, we provide a stringent test of model fidelity, establishing the effectiveness of our framework for real-world robotic tasks.

\section{Experiments}
\label{sec:experiments}

We validate \method through a comprehensive set of experiments across diverse robotic systems, environments, and network architectures.
The experiments are designed to assess the accuracy and robustness of \method, evaluate its architectural and training design choices, and demonstrate its effectiveness across diverse robotic tasks in Isaac Lab~\citep{mittal2023orbit} and in real-world deployment combined with \policymethod.
We start the analysis by looking into the autoregressive prediction accuracy and robustness of the world model learned with simulation data induced by a velocity tracking policy.
The observation and action spaces of the world model are detailed in \tabref{table:world_model_observation_space} and \tabref{table:action_space}.
We then compare various network architectures and the error induced across diverse robotic environments and tasks to demonstrate the generality of \method.
And finally, we learn a policy in \method with the proposed \policymethod and demonstrate the applicability and robustness of the method on ANYmal\,D~\citep{hutter2016anymal} and Unitree\,G1 hardware.

\subsection{Autoregressive Trajectory Prediction}
\begin{figure}
    \centering
    \begin{subfigure}{0.49\linewidth}
        \includegraphics[width=\linewidth]{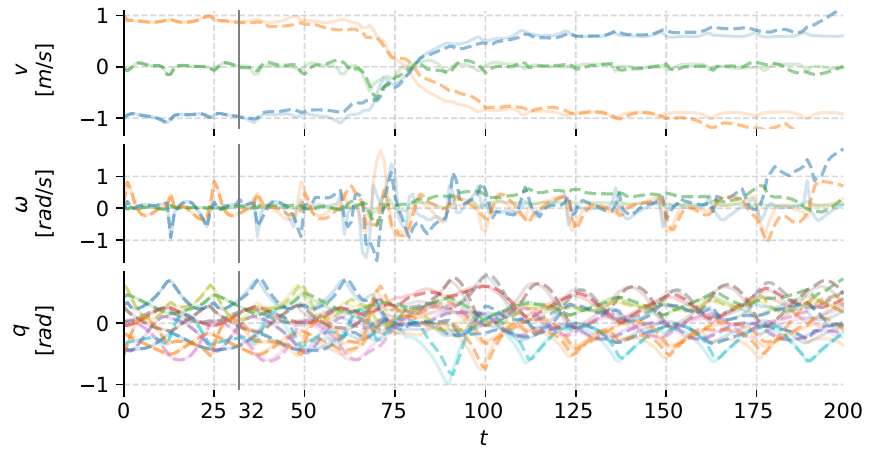}
        \caption{Autoregressive trajectory prediction by \method.}
        \label{fig:autoregressive_traj_pred}
    \end{subfigure}
    \begin{subfigure}{0.49\linewidth}
        \includegraphics[width=\linewidth]{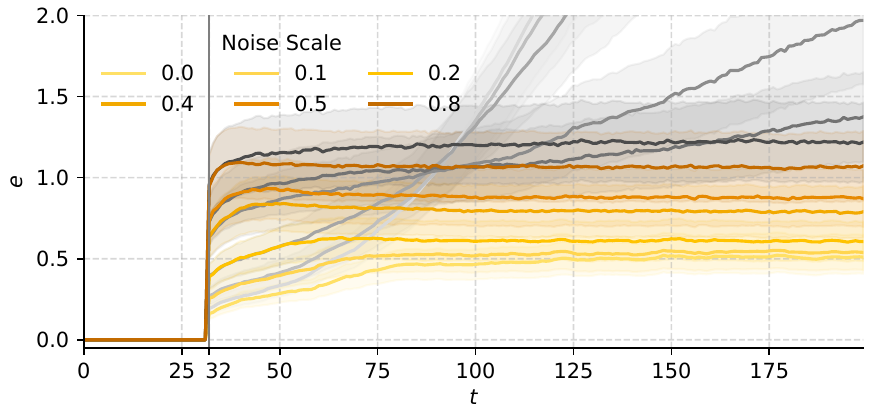}
        \caption{Prediction error under Gaussian noise.}
        \label{fig:autoregressive_traj_error_comparison}
    \end{subfigure}
    
    \caption{(Left) Solid lines represent ground truth trajectories, while dashed lines denote predicted state evolution. Predictions commence at $t = 32$ using historical observations, with future observations predicted autoregressively by feeding prior predictions back into the model. (Right) Yellow curves denote \method at varying noise levels, demonstrating consistent robustness and lower error accumulation across forecast steps. Grey curves represent the MLP baseline, which exhibits significantly higher error accumulation and reduced robustness to noise.}
    \label{fig:autoregressive_pred}
\end{figure}

The capability of a world model to maintain high fidelity during autoregressive rollouts is critical for effective planning and policy optimization.
To evaluate this aspect, we analyze the autoregressive prediction performance of \method using trajectories collected from ANYmal\,D hardware.
The control frequency of the robot is at $50\,Hz$.
The model is trained with history horizon $M = 32$ and forecast horizon $N = 8$.
Further details on the network architecture and training parameters are summarized in \secref{supp:robotic_world_model_architecture} and \secref{supp:robotic_world_model_training}, respectively.
The autoregressive trajectory predictions by \method are visualized in \figref{fig:autoregressive_traj_pred}.

% \begin{figure}
%     \centering
%     \includegraphics[width=\linewidth]{images/autoregressive_traj_pred.pdf}
%     \caption{Autoregressive trajectory prediction by \method. Solid lines represent ground truth trajectories, while dashed lines denote predicted state evolution. The control frequency of the robot is at $50\,Hz$. The model is trained with history horizon $M = 32$ and forecast horizon $N = 8$. Predictions commence at $t = 32$ using historical observations, with future observations predicted autoregressively by feeding prior predictions back into the model.}
%     \label{fig:autoregressive_traj_pred}
% \end{figure}

% \begin{figure}
%     \centering
%     \includegraphics[width=\linewidth]{images/autoregressive_traj_error_comparison.pdf}
%     \caption{Autoregressive trajectory prediction error under Gaussian noise. Yellow curves denote \method at varying noise levels, demonstrating consistent robustness and lower error accumulation across forecast steps. Grey curves represent the MLP baseline, which exhibits significantly higher error accumulation and reduced robustness to noise.}
%     \label{fig:autoregressive_traj_error_comparison}
% \end{figure}

% At time $t = 32$, the world model initiates predictions based on historical observations.
% From this point onward, future observations are predicted autoregressively using its prior outputs, simulating the deployment scenario in downstream applications.
% In \figref{fig:autoregressive_traj_pred}, the predicted observations are compared against the ground truth trajectories.

The results demonstrate that \method exhibits a remarkable alignment between predicted and ground truth trajectories across all observed variables.
This consistency persists over extended rollouts, showcasing the model’s ability to mitigate compounding errors—a critical challenge in long-horizon predictions.
% Despite the accumulated shifts at the far end of the prediction, the variables maintain close fidelity to the actual trajectories, with minimal deviations, even in stochastic and dynamically complex environments.
% The proposed GRU-based autoregressive mechanism demonstrates superior predictive accuracy.
% The historical context encoded in the hidden states allows \method to effectively capture partially observable dynamics, a capability evident in the accurate long-term predictions.
This performance is attributed to the dual-autoregressive mechanism introduced in \secref{sec:self-supervised_autoregressive_training}, which stabilizes predictions despite the short forecast horizon employed during training.
% This robust predictive capability has profound implications for downstream tasks.
% The ability of \method to maintain high fidelity over extended predictions ensures that policies trained within the learned world model generalize effectively to physical hardware, as further validated in \secref{sec:policy_learning_and_hardware_transfer}.
A comparison of state evolution between the \method prediction and the ground truth simulation is illustrated in \figref{fig:prediction_visualization} (bottom).
% Both environments are initialized in the same state and propagated using the same policy to ensure a fair evaluation of the world model's predictive accuracy.
The visualization highlights the ability of \method to maintain consistency in trajectory predictions over long horizons, even beyond the training forecast horizon.
% Notably, \method effectively mitigates these issues, preserving the integrity of the predicted trajectories.
This robustness is pivotal for stable policy learning and deployment, as discussed further in \secref{sec:policy_learning_and_hardware_transfer}.
% The results underscore the capability of \method to produce accurate long-horizon predictions, validating its applicability to real-world robotic tasks.

It is notable that the choice of history horizon $M$ and forecast horizon $N$ plays a critical role in the training and performance of \method.
Our ablation study in \secref{supp:dual-autoregressive_mechanism} reveals that, while extending both $M$ and $N$ improves accuracy, practical considerations of computational cost necessitate careful tuning of these hyperparameters to achieve optimal performance.

\subsection{Robustness under Noise}
A critical challenge in training world models is their ability to generalize under noisy conditions, particularly when predictions rely on autoregressive rollouts.
Even small deviations from the training distribution can cascade into untrained regions, causing the model to hallucinate future trajectories.
% These issues are magnified in autoregressive settings due to the compounding nature of prediction errors.
To assess the robustness of \method, we analyze its performance under Gaussian noise perturbations applied to both observations and actions.
We compare the results with an MLP-based baseline also trained autoregressively with the same history and forecast horizon, as shown in \figref{fig:autoregressive_traj_error_comparison}, where yellow curves denote the relative prediction error $e$ for \method, and grey curves represent the MLP baseline.

The results indicate a clear advantage of \method over the MLP baseline across all noise levels.
As forecast steps increase, the relative prediction error of the MLP model grows significantly, diverging more rapidly than \method.
In contrast, \method demonstrates superior stability, maintaining lower prediction errors even under high noise levels.
This robustness can be attributed to the dual-autoregressive mechanism introduced in \secref{sec:self-supervised_autoregressive_training}, which ensures stability in long-horizon predictions.
% The inner autoregression updates the GRU hidden states during the historical context window, while the outer autoregression feeds back predictions during the forecast horizon.
This design minimizes the accumulation of errors by continually refining the state representation toward long-term predictions, even in the presence of noisy inputs.
% By encoding historical observation-action sequences into its hidden states, the GRU model provides a richer representation of the environment dynamics compared to the MLP, which processes the data in a feedforward manner, leading to a rapid deterioration of performance.

\subsection{Generality across Robotic Environments}
\begin{figure}
    \centering
    \includegraphics[width=\linewidth]{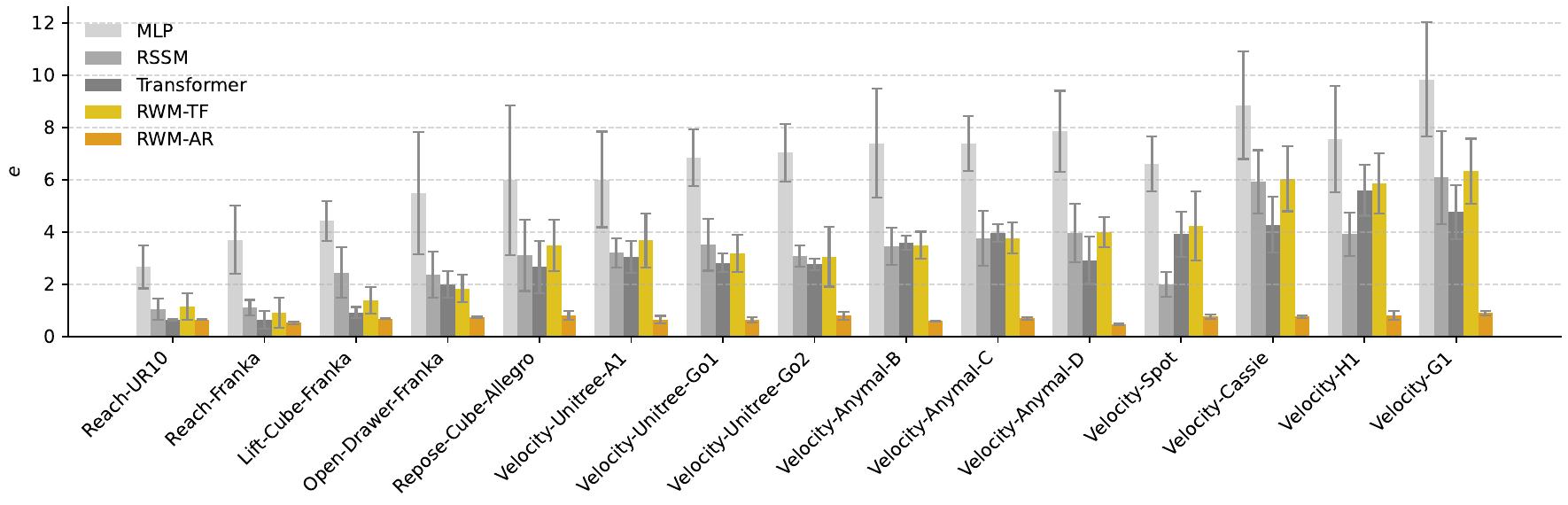}
    \caption{Autoregressive trajectory prediction errors across diverse robotic environments and network architectures. \method trained with autoregressive training (\method-AR) consistently outperforms baseline methods, including MLP, recurrent state-space model (RSSM), and transformer-based architectures. \method-AR demonstrates superior generalization and robustness across tasks, from manipulation to locomotion. Autoregressive training (\method-AR) reduces compounding errors over long rollouts, significantly improving performance compared to teacher-forcing training (\method-TF).}
    \label{fig:error_all_environments}
\end{figure}

To assess the generality and robustness of \method across a diverse range of robotic environments, we compare its performance with several baseline methods, including MLP, recurrent state-space model (RSSM)~\citep{hafner2019learning, hafner2019dream, hafner2020mastering, hafner2023mastering}, and transformer-based architectures~\citep{chen2021decision, reed2022generalist}.
These baselines represent widely adopted approaches in dynamics modeling and policy optimization.
All models are given the same context during training and evaluation.
Their training parameters are detailed in \secref{supp:baselines_architecture}.
The relative autoregressive prediction errors $e$ for these models are shown in \figref{fig:error_all_environments}.
The tasks span manipulation scenarios as well as quadruped and humanoid locomotion tasks, allowing for a comprehensive evaluation of the models.
In addition, we highlight the importance of the autoregressive training introduced in \secref{sec:self-supervised_autoregressive_training} by including both \method trained with teacher-forcing (\method-TF) and autoregressive training (\method-AR), demonstrating the significant performance gains achieved by the latter.

The results highlight the superiority of \method trained with autoregressive training (\method-AR), which consistently achieves the lowest prediction errors across all environments.
The performance gap between \method-AR and the baselines is especially pronounced in complex and dynamic tasks, such as velocity tracking for legged robots, where accurate long-horizon predictions are critical for effective control.
The comparison also reveals that \method-AR significantly outperforms its teacher-forcing counterpart (\method-TF), underscoring the importance of autoregressive training in mitigating compounding prediction errors over long rollouts.
We additionally visualize the imagination rolled out by \method-AR compared with the ground truth simulation in \figref{fig:prediction_visualization} and \figref{fig:prediction_visualization_more}.

Note that the baselines are trained using teacher forcing as they are traditionally implemented.
However, the proposed autoregressive training framework is architecture-agnostic and can also be applied to baseline models.
When trained with autoregressive training, RSSM achieves a performance comparable to the proposed GRU-based architecture.
Nevertheless, we opt for the GRU-based model due to its simplicity and computational efficiency.
On the other hand, training transformer architectures with autoregressive training does not scale effectively, as the multi-step gradient propagation in autoregressive forecasting leads to GPU memory constraints, limiting their practicality for this approach.
These results demonstrate that \method, when combined with autoregressive training, achieves robust and generalizable performance across diverse robotic tasks.
% Its ability to consistently outperform baseline architectures highlights the effectiveness of its design in addressing the challenges of long-horizon prediction and error accumulation.

\subsection{Policy Learning and Hardware Transfer}
\label{sec:policy_learning_and_hardware_transfer}

Using \policymethod, we train a goal-conditioned velocity tracking policy for ANYmal\,D and Unitree\,G1 leveraging \method.
% At the beginning of each imagination rollout, a target velocity is sampled alongside the initial observation retrieved from the replay buffer $\gD$.
% The policy selects an action based on the sampled observation and velocity command.
% \method then predicts the next observation based on the observation-action pair, and the reward is computed from this transition.
% This process continues recursively over the rollout horizon, with the collected imagination trajectories being used to update the policy.
% After each iteration, the updated policy interacts with the real environment, collecting additional data for the replay buffer, which is subsequently used to refine \method.
The policy's observation and action spaces are detailed in \secref{supp:observation_and_action_spaces}, and its architecture is described in \secref{supp:mbpo_ppo_architecture}.
Reward formulations are provided in \secref{supp:reward_functions}, while training parameters are summarized in \secref{supp:mbpo_ppo_training}.
% Note that learning policies with long imagination rollouts from a learned world model is especially challenging due to the inherent compounding errors in the autoregressive nature of the predictions arising from two key factors:
% The world model's reliance on previous predictions to infer future observations, which introduces drift over time.
% And the policy's dependence on predicted observations to compute actions, amplifying errors in the generated trajectories.
We compare \policymethod with two baselines: Short-Horizon Actor-Critic (SHAC)~\citep{xu2022accelerated} and DreamerV3~\citep{hafner2023mastering}.
SHAC employs a first-order gradient-based method that propagates gradients through the world model to optimize the policy.
Dreamer integrates a latent-space dynamics model with an actor-critic framework, emphasizing sample efficiency and robustness in continuous control tasks.
% The comparative results of model error and policy performance are shown in \figref{fig:mbpo_ppo_training}.

\begin{figure}
    \centering
    \includegraphics[width=\linewidth]{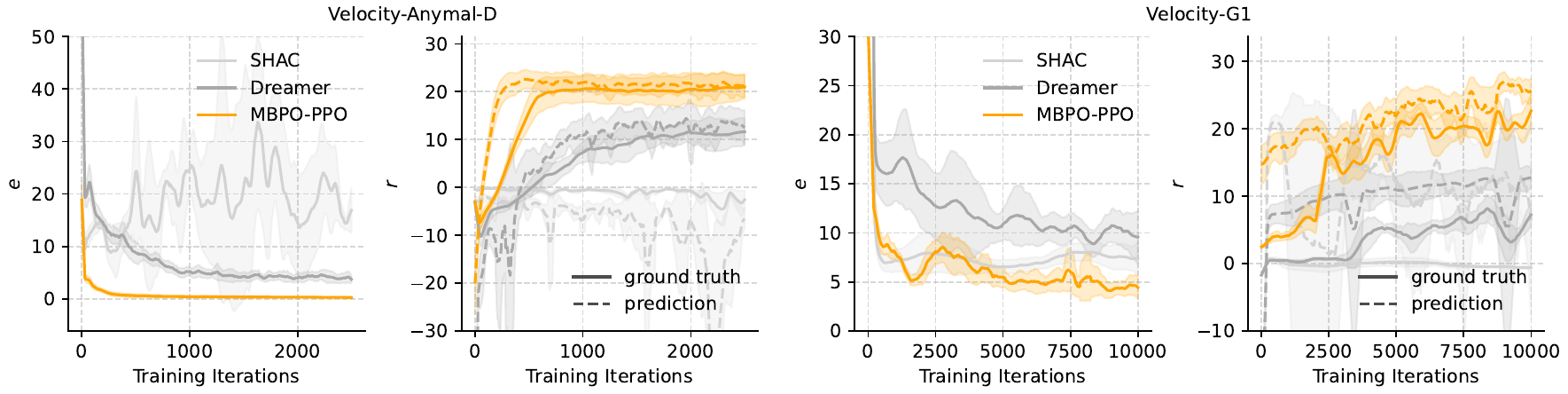}
    \caption{Model error and policy mean reward for the ANYmal\,D (left) and Unitree\,G1 (right) velocity tracking task with \policymethod. The policy is trained using estimated rewards computed from predicted observations by \method. Ground truth rewards, visualized with solid lines, are reported by the simulator for \textit{evaluation} purposes only.}
    \label{fig:mbpo_ppo_training}
\end{figure}

\Figref{fig:mbpo_ppo_training} illustrates the model error $e$ during policy optimization.
While \policymethod demonstrates a significant reduction in model error over training, SHAC struggles with high and fluctuating model error throughout the process.
% This instability arises from the poor quality of the data collected by the policy during training.
Its reliance on first-order gradients for optimization is not well-suited for discontinuous dynamics, such as those encountered in legged locomotion, where system behavior changes drastically due to varying contact patterns.
The resulting inaccurate gradients lead to suboptimal policy updates, producing chaotic robot behaviors during training.
These chaotic behaviors, in turn, generate low-quality training data for updating \method, exacerbating model inaccuracies.
% This feedback loop causes a compounding failure mode where both the policy and the model collapse.
% Dreamer exhibits lower model error than SHAC but still significantly higher error compared to \policymethod.
Although Dreamer effectively leverages its latent-space dynamics model for policy optimization, its reliance on shorter planning horizons during training limits its ability to handle long-horizon dependencies, particularly in stochastic environments.
As a result, Dreamer encounters moderate compounding errors during policy learning, which hinder its convergence to optimal behaviors.

On the right plot of rewards $r$, predicted rewards (dashed) from \policymethod initially overshoot the ground truth (solid) due to the policy exploiting small inaccuracies in the model’s optimistic estimates.
As training progresses, predictions align more closely with ground truth, remaining accurate enough to guide effective learning.
% Crucially, \policymethod achieves high final performance, indicating that early overestimation does not hinder convergence or deployment.
In contrast, SHAC fails to converge, producing unstable behaviors that degrade both policy and model quality.
Dreamer demonstrates partial convergence, achieving higher rewards compared to SHAC but significantly lagging behind \policymethod.

To evaluate the robustness of the learned policies, we deploy them on ANYmal\,D and Unitree\,G1 hardware in a zero-shot transfer setup.
SHAC and Dreamer fail to produce a deployable policy due to its collapse during training.
However, as shown in \figref{fig:prediction_visualization}, the policy learned using \policymethod demonstrates reliable and robust performance in tracking goal-conditioned velocity commands and maintaining stability under external disturbances, such as unexpected impacts and terrain conditions.
The success of \policymethod in hardware deployment is a direct result of the high-quality trajectory predictions generated by \method, which enable accurate and effective policy optimization.
% The results demonstrate that \method maintains sufficient accuracy to bridge the sim-to-real gap effectively.
% The final policies generalize well to real-world conditions, producing robust and stable behaviors in hardware deployment.
Videos showcasing the robustness of the policies on hardware, including their responses to external disturbances, are available on our webpage.
These results underline the effectiveness of \method and \policymethod in enabling robust and scalable policy deployment for real-world robotic systems.

\section{Limitations}
\label{sec:limitations}

The policy learned with \method and \policymethod surpasses existing MBRL methods in both robustness and generalization.
However, it still falls short of the performance achieved by well-tuned model-free RL methods trained on high-fidelity simulators.
Model-free RL, being a more mature and extensively optimized paradigm, excels in settings where unlimited interaction with near-perfect simulators is possible.
In contrast, the strengths of MBRL are more pronounced in scenarios where accurate or efficient simulation is infeasible, making it an indispensable tool for enabling intelligent agents to eventually learn and adapt in complex, real-world environments.
To clarify the computational and performance aspects, we provide a comparison against a PPO-based method with a high-fidelity simulator in \tabref{table:comparison_with_model-free_method}.
\begin{table}[h]
    \centering
    \vspace{-1em}
    \caption{Comparison with model-free method}
    \begin{tabular}{lcc|c}
    \toprule
        Method & \method pretraining & \policymethod & PPO \\
        \midrule
        state transitions & 6M & $-$ & 250M \\
        total training time & 50 min & 5 min & 10 min \\
        step inference time & $-$ & 1 ms & 1 ms \\
        real tracking reward & $-$ & $0.90 \pm 0.04$ & $0.90 \pm 0.03$ \\
    \bottomrule
    \end{tabular}
    \vspace{-1em}
    \label{table:comparison_with_model-free_method}
\end{table}

In this work, the world model is pre-trained using simulation data prior to policy optimization, reducing instability during training (see \secref{supp:collision_handling_and_model_pretraining}).
However, training from scratch remains challenging as policies can exploit model inaccuracies during exploration, leading to inefficiency and instability.
In addition, the need for additional interaction with the environment to fine-tune the world model highlights areas for further refinement.
Nevertheless, enabling safe and effective online learning directly on hardware remains challenging (see \secref{supp:challenges_in_real-world_online_learning}).
Current training in simulation avoids potential hardware damage, but incorporating safety constraints and robust uncertainty estimates will be critical for deploying \method and \policymethod in real-world, lifelong learning scenarios.
These limitations underscore the trade-offs inherent in MBRL frameworks, balancing data efficiency, safety, and performance while addressing the complexities of real-world robotic systems.

\section{Conclusion}
In this work, we present \method, a robust and scalable framework for learning world models tailored to complex robotic tasks.
Leveraging a dual-autoregressive mechanism, \method effectively addresses key challenges such as compounding errors, partial observability, and stochastic dynamics.
By incorporating historical context and self-supervised training over long prediction horizons, \method achieves superior accuracy and robustness without relying on domain-specific inductive biases, enabling generalization across diverse tasks.
Through extensive experiments, we demonstrate that \method consistently outperforms state-of-the-art approaches like RSSM and transformer-based architectures in autoregressive prediction accuracy across diverse robotic environments.
% Its ability to maintain stability under noise and dynamically complex scenarios further underscores its robustness and applicability to real-world tasks.
Building on \method, we propose \policymethod, a policy optimization framework that leverages long world model rollout fidelity.
Policies trained using \policymethod demonstrate superior performance in simulation and transfer seamlessly to hardware, as evidenced by zero-shot deployment on the ANYmal\,D and Unitree\,G1 robots.
% Unlike baseline methods such as SHAC and Dreamer, which struggle with unstable dynamics and compounding errors, \policymethod reliably optimizes policies that generalize effectively to physical systems.
This work advances the field of model-based reinforcement learning by providing a generalizable, efficient, and scalable framework for learning and deploying world models.
The results highlight \method’s potential to enable adaptive, robust, and high-performing robotic systems, setting a foundation for broader adoption of model-based approaches in real-world applications.
% To support future extensions and deployment, we also analyze potential limitations of our approach in \secref{supp:limitations}.

\begin{ack}
This research was supported by the ETH AI Center.
\end{ack}

\clearpage
\bibliographystyle{unsrt}
\bibliography{main}

%%%%%%%%%%%%%%%%%%%%%%%%%%%%%%%%%%%%%%%%%%%%%%%%%%%%%%%%%%%%

\clearpage
\appendix

\section{Technical Appendices and Supplementary Material}
%Figures and Tables will have S in the name
\renewcommand{\thetable}{S\arabic{table}}
\renewcommand{\thefigure}{S\arabic{figure}}
\renewcommand{\theequation}{S\arabic{equation}}

\subsection{Task Representation}
\label{supp:task_representation}

\subsubsection{Observation and action spaces}
\label{supp:observation_and_action_spaces}
The observation space for the ANYmal\,D and Unitree\,G1 world model is composed of base linear and angular velocities $v$, $\omega$ in the robot frame, measurement of the gravity vector in the robot frame $g$, joint positions $q$, velocities $\dot{q}$ and torques $\tau$ as in \tabref{table:world_model_observation_space}.

\begin{table}[h]
    \centering
    \caption{World model observation space}
    \begin{tabular}{lcc|lcc}
    \toprule
        Entry & Symbol & Dimensions & Entry & Symbol & Dimensions \\
        \midrule
        \textit{ANYmal\,D} & & & \textit{Unitree\,G1} & & \\
        base linear velocity & $v$ & 0:3 & base linear velocity & $v$ & 0:3 \\
        base angular velocity & $\omega$ & 3:6 & base angular velocity & $\omega$ & 3:6 \\
        projected gravity & $g$ & 6:9 & projected gravity & $g$ & 6:9 \\
        joint positions & $q$ & 9:21 & joint positions & $q$ & 9:38 \\
        joint velocities & $\dot{q}$ & 21:33 & joint velocities & $\dot{q}$ & 38:67 \\
        joint torques & $\tau$ & 33:45 & joint torques & $\tau$ & 67:96 \\
    \bottomrule
    \end{tabular}
    \label{table:world_model_observation_space}
\end{table}

The privileged information is used to provide an additional learning objective that implicitly embeds critical information for accurate long-term
predictions.
The space is composed of knee and foot contacts as in \tabref{table:world_model_privileged_information_space}.

\begin{table}[h]
    \centering
    \caption{World model privileged information space}
    \begin{tabular}{lcc|lcc}
    \toprule
        Entry & Symbol & Dimensions & Entry & Symbol & Dimensions \\
        \midrule
        \textit{ANYmal\,D} & & & \textit{Unitree\,G1} & & \\
        knee contact & $-$ & 0:4 & body contact & $-$ & 0:26 \\
        foot contact & $-$ & 4:8 & foot height & $-$ & 26:28 \\
        & & & foot velocity & $-$ & 28:30 \\
    \bottomrule
    \end{tabular}
    \label{table:world_model_privileged_information_space}
\end{table}

The action space is composed of joint position targets as in \tabref{table:action_space}.

\begin{table}[h]
    \centering
    \caption{Action space}
    \begin{tabular}{lcc|lcc}
    \toprule
        Entry & Symbol & Dimensions & Entry & Symbol & Dimensions \\
        \midrule
        \textit{ANYmal\,D} & & & \textit{Unitree\,G1} & & \\
        joint position targets & $q^*$ & 0:12 & joint position targets & $q^*$ & 0:29 \\
    \bottomrule
    \end{tabular}
    \label{table:action_space}
\end{table}

The observation space for the ANYmal velocity tracking policy is composed of base linear and angular velocities $v$, $\omega$ in the robot frame, measurement of the gravity vector in the robot frame $g$, velocity command $c$, joint positions $q$ and velocities $\dot{q}$ as in \tabref{table:policy_observation_space}.

\begin{table}[h]
    \centering
    \caption{Policy observation space}
    \begin{tabular}{lcc|lcc}
    \toprule
        Entry & Symbol & Dimensions & Entry & Symbol & Dimensions \\
        \midrule
        \textit{ANYmal\,D} & & & \textit{Unitree\,G1} & & \\
        base linear velocity & $v$ & 0:3 & base linear velocity & $v$ & 0:3 \\
        base angular velocity & $\omega$ & 3:6 & base angular velocity & $\omega$ & 3:6 \\
        projected gravity & $g$ & 6:9 & projected gravity & $g$ & 6:9 \\
        velocity command & $c$ & 9:12 & velocity command & $c$ & 9:12 \\
        joint positions & $q$ & 12:24 & joint positions & $q$ & 12:41 \\
        joint velocities & $\dot{q}$ & 24:36 & joint velocities & $\dot{q}$ & 41:70 \\
        last actions & $a'$ & 36:48 & last actions & $a'$ & 70:99 \\
    \bottomrule
    \end{tabular}
    \label{table:policy_observation_space}
\end{table}

\subsubsection{Reward functions}
\label{supp:reward_functions}

The total reward is sum of the following terms with weights detailed in \tabref{table:reward_weights}.

\begin{table}[h]
    \centering
    \caption{Reward weights}
    \begin{tabular}{llll|llll}
    \toprule
        Symbol & Value & Symbol & Value & Symbol & Value & Symbol & Value \\
        \midrule
        \textit{ANYmal\,D} & & & & \textit{Unitree\,G1} & & & \\
        $w_{v_{xy}}$ & $1.0$ & $w_{\omega_z}$ & $0.5$ & $w_{v_{xy}}$ & $1.0$ & $w_{\omega_z}$ & $0.5$ \\
        $w_{v_z}$ & $-2.0$ & $w_{\omega_{xy}}$ & $-0.05$ & $w_{v_z}$ & $-2.0$ & $w_{\omega_{xy}}$ & $-0.05$ \\
        $w_{q_\tau}$ & $-2.5e^{-5}$ & $w_{\ddot{q}}$ & $-2.5e^{-7}$ & $w_{q_\tau}$ & $-2.5e^{-5}$ & $w_{\ddot{q}}$ & $-2.5e^{-7}$ \\
        $w_{\dot{a}}$ & $-0.01$ & $w_{f_a}$ & $0.5$ & $w_{\dot{a}}$ & $-0.05$ & $w_{f_a}$ & $0.0$ \\
        $w_c$ & $-1.0$ & $w_g$ & $-5.0$ & $w_c$ & $-1.0$ & $w_g$ & $-5.0$ \\
        $w_{f_c}$ & $0.0$ & $w_{q_d}$ & $0.0$ & $w_{f_c}$ & $1.0$ & $w_{q_d}$ & $-1.0$ \\
        
    \bottomrule
    \end{tabular}
    \label{table:reward_weights}
\end{table}

\subsubsubsection{Linear velocity tracking $x,y$}

\begin{equation*}
    r_{v_{xy}} = w_{v_{xy}} e^{-\| c_{xy} - v_{xy} \|_2^2 / \sigma_{v_{xy}}^2},
\end{equation*}
where $\sigma_{v_{xy}} = 0.25$ denotes a temperature factor, $c_{xy}$ and $v_{xy}$ denote the commanded and current base linear velocity.

\subsubsubsection{Angular velocity tracking}

\begin{equation*}
    r_{\omega_z} = w_{\omega_z} e^{- \| c_z - \omega_z \|_2^2 / \sigma_{\omega_z}^2},
\end{equation*}
where $\sigma_{\omega_z} = 0.25$ denotes a temperature factor, $c_z$ and $\omega_z$ denote the commanded and current base angular velocity.

\subsubsubsection{Linear velocity $z$}

\begin{equation*}
    r_{v_z} = w_{v_z} \left \| v_z \right \|_2^2,
\end{equation*}
where $v_z$ denotes the base vertical velocity.

\subsubsubsection{Angular velocity $x,y$}

\begin{equation*}
    r_{\omega_{xy}} = w_{\omega_{xy}} \left \| \omega_{xy} \right \|_2^2,
\end{equation*}
where $\omega_{xy}$ denotes the current base roll and pitch velocity.

\subsubsubsection{Joint torque}

\begin{equation*}
    r_{q_\tau} = w_{q_\tau} \left \| \tau \right \|_2^2,
\end{equation*}
where $\tau$ denotes the joint torques.

\subsubsubsection{Joint acceleration}

\begin{equation*}
    r_{\ddot{q}} = w_{\ddot{q}} \left \| \ddot{q} \right \|_2^2,
\end{equation*}
where $\ddot{q}$ denotes the joint acceleration.

\subsubsubsection{Action rate}

\begin{equation*}
    r_{\dot{a}} = w_{\dot{a}} \| a' - a \|_2^2,
\end{equation*}
where $a'$ and $a$ denote the previous and current actions.

\subsubsubsection{Feet air time}

\begin{equation*}
    r_{f_a} = w_{f_a} t_{f_a},
\end{equation*}
where $t_{f_a}$ denotes the sum of the time for which the feet are in the air.

\subsubsubsection{Undesired contacts}

\begin{equation*}
    r_c = w_c c_u,
\end{equation*}
where $c_u$ denotes the counts of the undesired contacts.

\subsubsubsection{Flat orientation}

\begin{equation*}
    r_g = w_g g_{xy}^2,
\end{equation*}
where $g_{xy}$ denotes the $xy$-components of the projected gravity.

\subsubsubsection{Foot clearance}

\begin{equation*}
    r_{f_c} = w_{f_c} h_{f_c},
\end{equation*}
where $h_{f_c}$ denotes the clearance height of the swing feet.

\subsubsubsection{Joint deviation}

\begin{equation*}
    r_{q_d} = w_{q_d} \left \| q - q_0 \right \|_1,
\end{equation*}
where $q_0$ denotes the default joint position.

\subsection{Network Architecture}
\label{supp:network_architecture}

\subsubsection{\method}
\label{supp:robotic_world_model_architecture}
\begin{figure}
    \centering
    \includegraphics[width=0.4\linewidth]{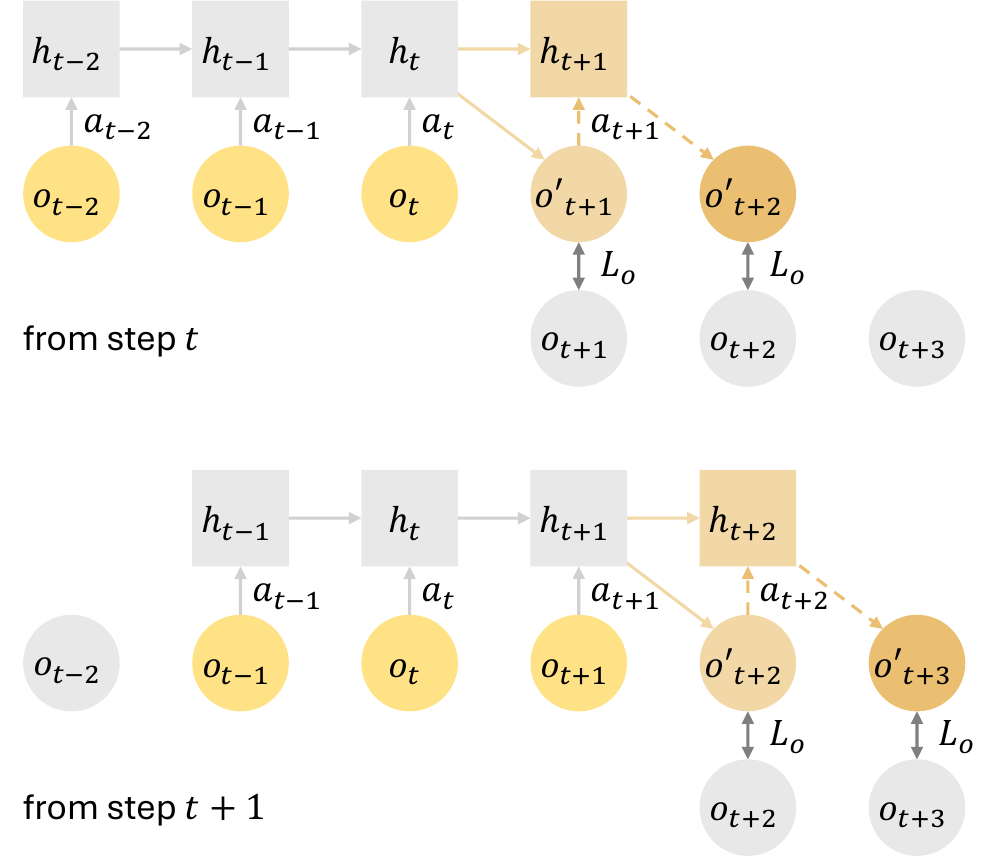}
    \caption{Dual-autoregressive mechanism employed in \method. Inner autoregression updates GRU hidden states after each historical step within the context horizon, while outer autoregression feeds predicted observations from the forecast horizon back into the network. The dashed arrows denote the sequential autoregressive prediction steps, highlighting robustness to long-term dependencies and transitions.}
    \label{fig:dual-autoregressive}
\end{figure}

The robotic world model consists of a GRU base and MLP heads predicting the mean and standard deviation of the next observation and privileged information such as contacts, as detailed in \tabref{table:robotic_world_model_architecture}.
The training scheme is visualized in \figref{fig:dual-autoregressive}.

\begin{table}[h]
    \centering
    \caption{\method architecture}
    \begin{tabular}{lccc}
    \toprule
        Component & Type & Hidden Shape & Activation \\
        \midrule
        base & GRU & 256, 256 & $-$ \\
        heads & MLP & 128 & ReLU \\
    \bottomrule
    \end{tabular}
    \label{table:robotic_world_model_architecture}
\end{table}

\subsubsection{Baselines}
\label{supp:baselines_architecture}
The network architectures of the baselines are detailed in \tabref{table:baseline_architecture}.

\begin{table}[h]
    \centering
    \caption{Baseline architecture}
    \begin{tabular}{llc}
    \toprule
        Network & Parameter & Value \\
        \midrule
        MLP & hidden shape & 256, 256 \\
        & activation & ReLU \\
        \midrule
        RSSM & type & GRU \\
        & hidden size & 256 \\
        & layers & 2 \\
        & latent dimension & 64 \\
        & prior type & categorical \\
        & categories & 32 \\
        \midrule
        Transformer & type & decoder \\
        & dimension & 64 \\
        & heads & 8 \\
        & layers & 2 \\
        & context length & 32 \\
        & positional encoding & sinusoidal \\
    \bottomrule
    \end{tabular}
    \label{table:baseline_architecture}
\end{table}

\subsubsection{\policymethod}
\label{supp:mbpo_ppo_architecture}
\begin{figure}
    \centering
    \includegraphics[width=0.4\linewidth]{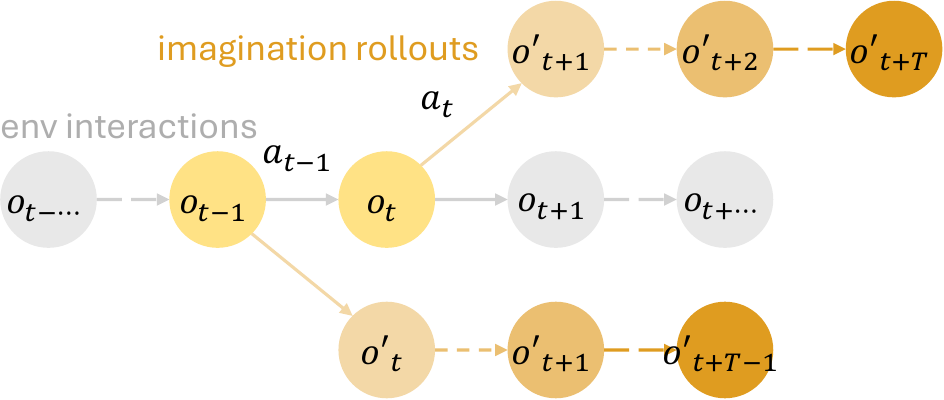}
    \caption{Model-Based Policy Optimization with learned world models. The framework combines real environment interactions with simulated rollouts for efficient policy optimization. Observation and action pairs from the environment are stored in a replay buffer and used to train the autoregressive world model. Imagination rollouts using the learned model predict future states over a horizon of $T$, providing trajectories for policy updates through reinforcement learning algorithms.}
    \label{fig:mbpo}
\end{figure}

The network architectures of the policy and the value function used in \policymethod are detailed in \tabref{table:policy_and_value_function_architecture}.
The training scheme is visualized in \figref{fig:mbpo}.

\begin{table}[h]
    \centering
    \caption{Policy and value function architecture}
    \begin{tabular}{lccc}
    \toprule
        Network & Type & Hidden Shape & Activation \\
        \midrule
        policy & MLP & 128, 128, 128 & ELU \\
        value function & MLP & 128, 128, 128 & ELU \\
    \bottomrule
    \end{tabular}
    \label{table:policy_and_value_function_architecture}
\end{table}

\subsection{Training Parameters}
\label{supp:training_parameters}
The learning networks and algorithm are implemented in PyTorch 2.4.0 with CUDA 12.6 and trained on an NVIDIA RTX 4090 GPU.

\subsubsection{\method}
\label{supp:robotic_world_model_training}
The training information of \method is summarized in \tabref{table:robotic_world_model_training}.

\begin{table}[h]
\centering
    \caption{\method training parameters}
    \begin{tabular}{lcc}
    \toprule
        Parameter & Symbol & Value \\
        \midrule
        step time seconds & $\Delta t$ & $0.02$ \\
        max iterations & $-$ & $2500$ \\
        learning rate & $-$ & $1e^{-4}$ \\
        weight decay & $-$ & $1e^{-5}$ \\
        batch size & $-$ & $1024$ \\
        history horizon & $M$ & $32$ \\
        forecast horizon & $N$ & $8$ \\
        forecast decay & $\alpha$ & $1.0$ \\
        approximate training hours & $-$ & $1$ \\
        number of seeds & $-$ & $5$ \\
    \bottomrule
    \end{tabular}
    \label{table:robotic_world_model_training}
\end{table}

\subsubsection{\policymethod}
\label{supp:mbpo_ppo_training}

The training information of \policymethod is summarized in \tabref{table:policy_and_value_function_training}.

\begin{table}[h]
\centering
    \caption{\policymethod training parameters}
    \begin{tabular}{lcc}
    \toprule
        Parameter & Symbol & Value \\
        \midrule
        imagination environments & $-$ & $4096$ \\
        imagination steps per iteration & $-$ & $100$ \\
        step time seconds & $\Delta t$ & $0.02$ \\
        buffer size & $|\gD|$ & $1000$ \\
        max iterations & $-$ & $2500$ \\
        learning rate & $-$ & $0.001$ \\
        weight decay & $-$ & $0.0$ \\
        learning epochs & $-$ & $5$ \\
        mini-batches & $-$ & $4$ \\
        KL divergence target & $-$ & $0.01$ \\
        discount factor & $\gamma$ & $0.99$ \\
        clip range & $\epsilon$ & $0.2$ \\
        entropy coefficient & $-$ & $0.005$ \\
        number of seeds & $-$ & $5$ \\
    \bottomrule
    \end{tabular}
    \label{table:policy_and_value_function_training}
\end{table}

\subsection{Additional Experiments and Discussions}
\label{supp:additional_experiments_and_discussions}

\subsubsection{Dual-autoregressive Mechanism}
\label{supp:dual-autoregressive_mechanism}

\begin{figure}
    \centering
    \includegraphics[width=0.5\linewidth]{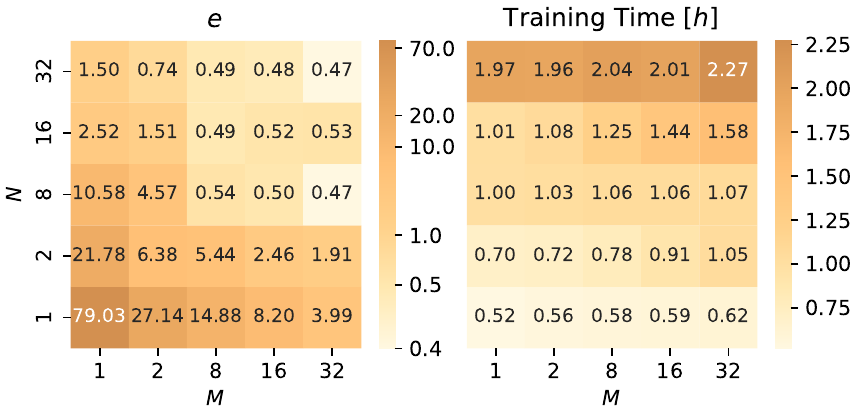}
    \caption{Ablation study on the history horizon $M$ and forecast horizon $N$ in \method. The heatmap on the left shows the relative autoregressive prediction error, with darker colors indicating higher errors. Models trained with larger history horizons $M$ exhibit lower errors, although the improvements plateau beyond a certain point. Forecast horizon $N$ has a significant impact, with longer horizons leading to better long-term prediction accuracy due to exposure to extended rollouts during training. The heatmap on the right illustrates training time, with darker colors representing longer durations. Increasing $N$ significantly raises training time due to sequential computation, while shorter horizons (e.g., $N=1$, teacher-forcing) enable faster training but result in poor prediction accuracy.}
    \label{fig:horizon_ablation}
\end{figure}

The heatmap on the left in \figref{fig:horizon_ablation} shows the relative autoregressive prediction error $e$ under different combinations of $M$ and $N$.
Models trained with a longer history horizon $M$ consistently exhibit lower prediction errors, demonstrating the importance of providing sufficient historical context to capture the underlying dynamics.
However, the influence of $M$ plateaus beyond a certain point, indicating diminishing returns for very large history horizons.
Forecast horizon $N$, on the other hand, plays a decisive role in improving long-term prediction accuracy.
Increasing $N$ during training leads to better performance in autoregressive rollouts, as it encourages the model to learn representations robust to compounding errors over extended prediction horizons.
This improvement comes at the cost of increased training time, as shown in the heatmap on the right.
Larger $N$ values require sequential computation during training due to the autoregressive nature of the process, significantly lengthening the training duration.

Interestingly, when the forecast horizon $N = 1$ (teacher-forcing), training can be highly parallelized, resulting in minimal training time.
However, this setting leads to poor autoregressive performance, as the model lacks exposure to long-horizon prediction during training and fails to effectively handle compounding errors.
From the results, an optimal trade-off emerges: moderate values of $M$ and $N$ balance prediction accuracy and training efficiency.
For instance, a history horizon of $M = 32$ and forecast horizon of $N = 8$ achieve strong autoregressive performance with manageable training time.
These settings ensure sufficient historical context while training the model for robust long-term predictions.
Overall, the results highlight the critical interplay between history and forecast horizons in autoregressive training.
While extending both $M$ and $N$ improves accuracy, practical considerations of computational cost necessitate careful tuning of these hyperparameters to achieve optimal performance.

\subsubsection{Visualization of Imagination Rollouts}

\begin{figure}
    \centering
    \includegraphics[width=\linewidth]{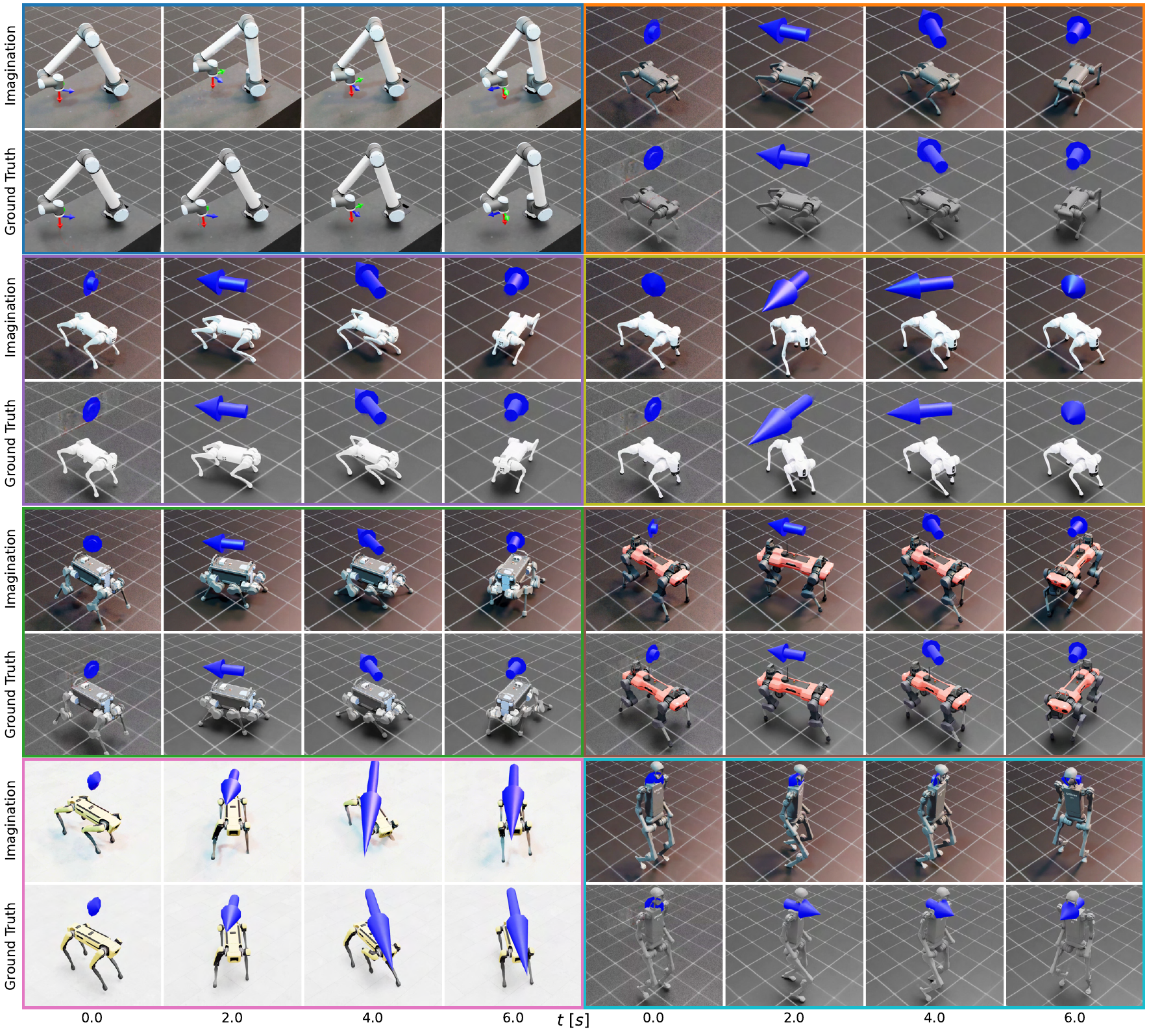}
    \caption{Autoregressive imagination of \method and ground-truth simulation across diverse robotic systems. For each environment, the top row showcases the \method autoregressively predicting future trajectories in imagination. The second row visualizes the ground truth evolution in simulation. The visualized coordinate and arrow markers denote the predicted and measured end-effector pose and base velocity, respectively.}
    \label{fig:prediction_visualization_more}
\end{figure}

The imagination rollouts across various robotic environments compared with the ground-truth simulation is visualized in \figref{fig:prediction_visualization_more}.

\subsubsection{Collision Handling and Model Pretraining}
\label{supp:collision_handling_and_model_pretraining}

In both phases of the pretraining and online fine-tuning of \method, we terminate rollouts and reset the environment when ground contact by the base is detected, signaling a failure.
We explicitly train \method to predict such terminations in its privileged information prediction head.
This enables the world model to learn transitions leading to unsafe situations.
During policy optimization, \policymethod treats these termination predictions as episode-ending events in imagination rollouts, affecting PPO’s return computation and state values.

\method is pretrained with simulation data induced by policies trained for similar tasks under varied dynamics.
The policy is learned from scratch purely in imagination, with \method fine-tuned using a \textit{single}-environment online dataset.
Pretraining is essential for two key reasons.
First, the online dataset is extremely limited, as it is generated by only a \textit{single} environment, akin to real-world constraints.
Training the world model entirely from scratch on such data would lead to severe overfitting and long training times.
Second, an immature policy would frequently cause the robot to fall, generating transitions with limited value.
In cases of significant failure or domain shift, training the world model solely on these data would result in chaotic imagined rollouts, which in turn would produce poor policy updates.
Pretraining stabilizes training and serves as a robust initialization for online fine-tuning, particularly in environments with challenging dynamics.

Importantly, \method pretraining does not require data from optimal policies.
\Figref{fig:autoregressive_pred} demonstrate that \method remains robust to domain shifts and injected noise.
As an alternative, we warm up the model using data from a suboptimal policy, which significantly stabilizes training.
Notably, this pretraining is only necessary for locomotion tasks due to the discontinuous dynamics and environment terminations.
Our manipulation experiments do not require such pretraining.

\subsubsection{Challenges in Real-World Online Learning}
\label{supp:challenges_in_real-world_online_learning}

We acknowledge that the advantages of our approach would be further demonstrated by performing the policy training phase directly on real hardware.
While this is a key long-term objective, several challenges currently prevent real-world deployment.

During online learning, the policy often exploits minor world model errors, leading to overly optimistic behaviors that result in collisions.
In simulation, these failures serve as corrective signals, but in real hardware, they pose a risk to the robot.
Our experiments show that such failures occur more than 20 times on average during online learning, which would be detrimental to real-world systems.
Even if hardware collisions were acceptable, fully automating online learning would require a recovery policy capable of resetting the robot to an initial state—a particularly challenging requirement for large platforms like ANYmal\,D or Unitree\,G1.
Additionally, privileged information used to fine-tune \method (e.g., contact forces) must be either measured or estimated using onboard sensors, which may not always be available.
To mitigate error exploitation, uncertainty-aware world models could be explored, but integrating such models into \method would require additional architectural modifications.
Due to these challenges, we approximate real-world constraints by using only a \textit{single} simulation environment with domain shifts from pretraining environments.
This setup reduces engineering effort while proving the feasibility of our approach. Our ongoing work specifically addresses these issues.

\subsection{Ethics and Societal Impacts}
\label{supp:ethics_and_societal_impacts}

This work does not involve human subjects or sensitive data.
All experiments are conducted in simulation or on dedicated robotic hardware operated by the authors, with no use of third-party datasets.
The research complies with the Code of Ethics of the venue.
The proposed framework provides a robust and scalable method for learning world models tailored to complex robotic tasks.
This can benefit domains such as healthcare, disaster response, and logistics, and reduce environmental and hardware costs associated with physical experimentation.
Potential risks include misuse of the method in surveillance or autonomous enforcement systems, and the acceleration of automation in labor-sensitive sectors.
While such uses are not intended or explored in this work, the authors acknowledge the dual-use potential of generalizable control methods.
To mitigate safety risks, policy training occurs entirely in simulation, and deployment is limited to policies validated under domain shifts.
Failure events are explicitly modeled and used to terminate unsafe rollouts.
Online learning on hardware is deferred due to safety concerns and the absence of reliable recovery strategies.
Future work will explore uncertainty-aware models and safer online adaptation.

\end{document}